\documentclass{llncs}
\usepackage{llncsdoc}
\usepackage{graphicx}
\usepackage[ruled]{algorithm2e}
\usepackage{dsfont}
\usepackage{multirow}
\usepackage{booktabs}
\usepackage[fleqn]{amsmath}
\usepackage{mathtools}
\usepackage{leqno}
\newcommand{\argmax}[1]{\underset{#1}{\operatorname{argmax}}}

\begin{document}
\title{Co-Multistage of Multiple Classifiers for Imbalanced Multiclass Learning}

\author{ }

\institute{ } 

\maketitle
\begin{abstract}
In this work, we propose two stochastic architectural models (\textit{CMC} and \textit{CMC-M}) with two layers of classifiers applicable to datasets with one and multiple skewed classes. This distinction becomes important when the datasets have a large number of classes.
Therefore, we present a novel solution to imbalanced multiclass learning with several skewed majority classes, which improves minority classes identification. This fact is particularly important for text classification tasks, such as event detection. Our models combined with preprocessing sampling techniques improved the classification results on 6 well-known datasets. Finally, we have also introduced a new metric \textit{SG-Mean} to overcome the multiplication by zero limitation of \textit{G-Mean}.

\end{abstract}
\section{Introduction}
Many real-world classification tasks, such as email analysis~\cite{Cui:2005,Tseng:2007,Zhou:Pei:2009}, 
fraud detection~\cite{Cao:2008}, 
medical diagnostics~\cite{Sheng:2006,Lin:2008}, face recognition,
discrimination aware classification~\cite{Bailey:2013},
 text classification~\cite{Tsukada:2001,Yu:2005,Haixun:2006}, and species distribution~\cite{Johnson:2012} can have a highly skewed or imbalanced class distributions datasets. Theoretically, a dataset is defined to have a imbalanced distribution when at least one class has an unequal number of instances relative to others. However, the community restricts the definition of imbalanced to datasets showing high or extreme imbalanced rates. High imbalanced rates are the overarching issue for supervised machine learning algorithms.
The first reported work focused on the multiclass imbalance problem~\cite{sun2006boosting},  extended the \textit{G-mean} metric~\cite{kubat1997addressing} to multiclass problem.
 While previous work~\cite{abe2004iterative,zhou2006multi} on cost-sensitive learning for multiclass settings addresses the problem of multiclass imbalance, the cost matrix was already available or manually created for a two-class scenario based on the nature of the problem. This manual definition does not scale well to the multiple-class case, where manually finding cost values is a hard task.

Recently, a review of imbalanced class problems~\cite{galar2012review} proposed a taxonomy for ensemble approaches to imbalanced binary class classification. Here, it is generalized to include multiclass classification and data pre-processing. As a result, it identifies three broad methodologies used to approach class imbalanced learning: Data Processing, Cost-Sensitive, and Hybrid. 
The \textbf{data preprocessing techniques} explored include resampling techniques to balance the class distribution. Among the resampling techniques are undersampling methods (e.g.: random undersampling), oversampling (e.g: random oversampling, \textit{SMOTE}~\cite{chawla2002smote}, and \textit{SMOTE} variations such as MSMOTE~\cite{hu2009msmote}, BorderLine-SMOTE~\cite{han2005borderline}), and a combination of both (e.g.: SPIDER~\cite{stefanowski2008selective}) 
AdaCost~\cite{fan1999adacost}, \textit{AdaC1,2,3}~\cite{sun2006boosting,sun2007cost} are examples of cost-sensitive learning based on boosting. These methods are the state-of-art multiclass imbalanced learning. There are also cost-sensitive rule base systems~\cite{Nguyen:2005}.
The \textbf{Hybrid category} includes a combination of Boosting, Bagging, or both, with one or more data preprocessing techniques, e.g.: SMOTEBoost~\cite{chawla2003smoteboost}, SMOTEBagging~\cite{Wang2009Diversity}, 
 EasyEnsemble~\cite{liu2009exploratory}, and BalanceCascade~\cite{liu2009exploratory}.

The contributions of this work include two new models to deal with two different distributive classes topologies and a new evaluation metric. Both topologies have imbalanced multiclass data distribution. The number of imbalanced classes defines the difference between them. While CMC model assumes only one imbalance class, the CMC-M model is dedicated for data with several imbalance classes.  
With the second stochastic model, we improved imbalanced multiclass classification of datasets with multiple skewed majority classes. Since it extends the first model, it was called \textit{Co-Multistage of Multiple Classifiers for Multiple Skewed Classes} (\textit{CMC-M}). It combines 1 binary view with 3 multiclass views of the data using 4 multistage ensemble classifiers~\cite{pudil:1992multistage}.

In this document, the next section 
describes the \textit{CMC}; Section \ref{section:CMC-M} presents the \textit{CMC-M}; Section \ref{sec:Experimental_Setup} details the experimental setup with the respective results. The conclusions end the document.

%

\section{\textit{CMC}: \textit{C}o-Multistage of \textit{M}ultiple \textit{C}lassifiers for Imbalanced Multiclass Learning}
\label{section:CMC}
\begin{figure}[h!]
\centering
\includegraphics[width=0.40\textwidth]{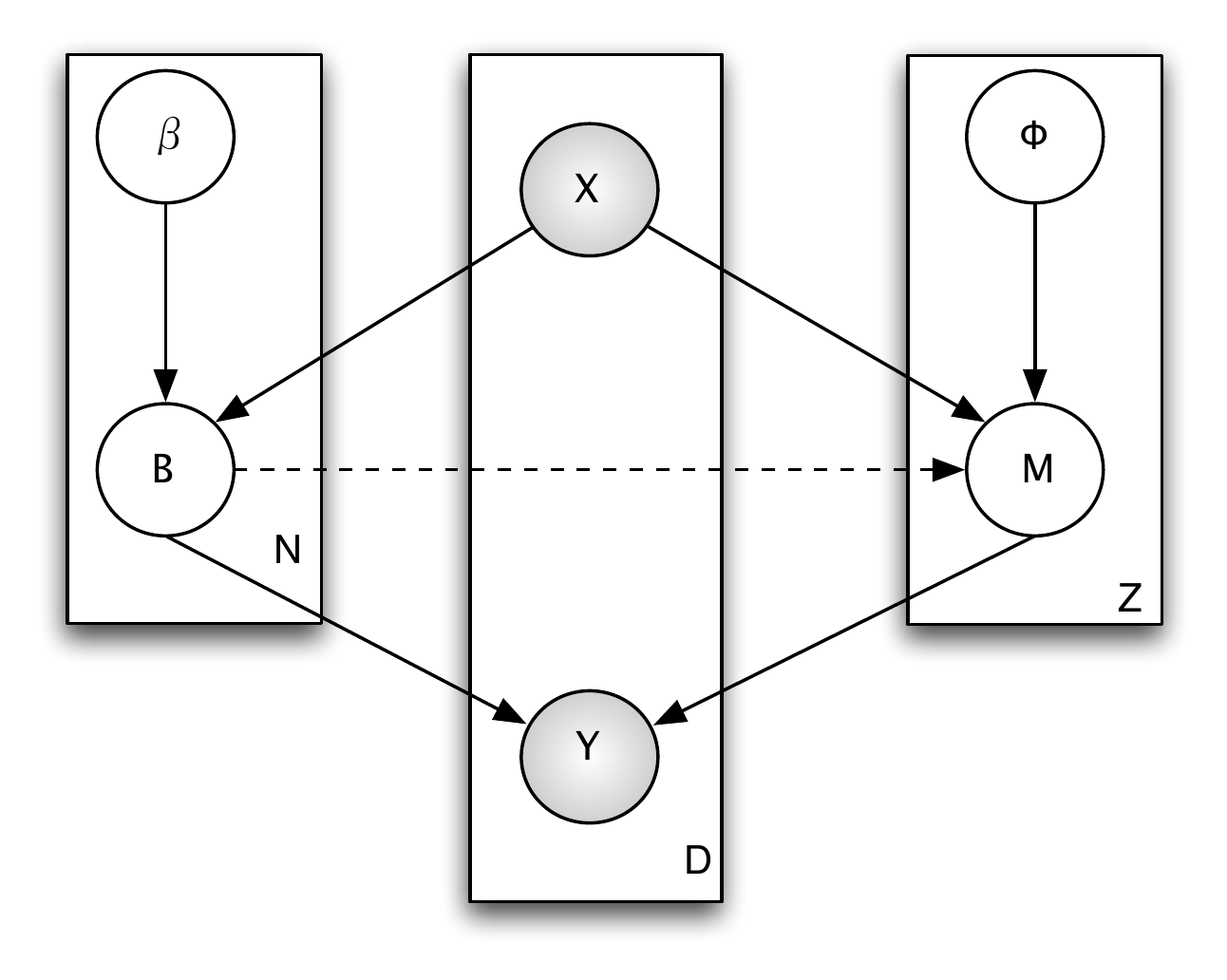}
\caption{Graphical model of CMC. In the plate notation, the boxes represent replication. The dashed arrow is an extension to the plate notation. It represents the prevalence of the parent variable distribution over the child under an a priori condition.}
\label{fig:GM_OneSkewedClass}
\end{figure}

\textit{Co-Multistage of Multiple Classifiers} (CMC) is a new stochastic architectural model composed of 2 state-layers: binary and multiclass classification.  Each layer has one multistage of multiple Classifiers (see Figure \ref{fig:GM_OneSkewedClass}). Both layers have 
latent variables $\beta$ and $\phi$ over classifiers to determine their activation in the multistage. We organized the layers in terms of complexity, i.e.: the first layer is the binary classifier which relaxes the problem to identify the majority class and then the second layer is multiclass classifier to identify primary the minority classes. 
As a result, the first layer is similar to the cascade learning framework for phishing detection~\cite{Xiang:2013}, which is a binary classification problem. While the second layer is an extension of the cascade framework to multiclass classification problems.

Before applying the \textit{CMC} method, it is necessary to generate two representations of the data in two label spaces (binary and multiclass). One keeps the original multiclass label space, while the other is binary. In the binary space (majority class vs. minority classes), all minority classes instances are relabeled to the minority class cluster label. 

These two data representations are used to train two multistage classifiers. Each multistage classifier is an ensemble of individual classifiers~\cite{pudil:1992multistage,Karypis:2006}. At every stage, the classifier confidence or probability of the most likely class is compared with the latent variable threshold. If it is greater or equal, the process stops. Otherwise, the classifier at the next stage is applied to the instance.
During training, all classifiers from both multistage classifiers can be trained in parallel. However, at classification time the binary multistage classifier is called first to decide whether it is a majority class. If the binary classifier is confident that it is a majority class, the classification stops, otherwise it continues to the multi-class multistage classifier.

\textit{CMC} assumes the following generative process for each label Y  in a set of instances D and described by a set of features X:
\begin{enumerate}

\item Choose $\beta_i\ \in\ ]0,1]$, where $i\ \in\ \{1,...,N\}$ and N is the number of stages in the binary classification (default value is 1). 

\item Choose $\Phi_i \in\ ]0,1]$, where $i\ \in\ \{1,...,Z\}$ and Z is the number of stages in the multiclass classification (default value is 1). 

\item For each label $Y_l$, where $l\ \in\ \{1,...,D\}$ and D is the number of instances.

	\begin{enumerate}
	\item Choose $B_{i,l}$ according to Equation \ref{eq:bin}; (b) Choose $M_{i,l}$ according to Equation \ref{eq:MultiClassOneSkew}; 
	(c) Choose a label $Y_{l}$ to Equation \ref{eq:mainOneSkew}
	\end{enumerate}
\end{enumerate}
\begin{equation}
\label{eq:bin}
P(B|X,\beta) = P(B_{i,l}|X_l) \sim Bernoulli(\beta_i) 
\end{equation}
\vspace{-0.5cm}
\begin{alignat*}{2}
    \text{where } & P(B_{i,l}|X_l) \ge \beta_i \ \wedge &  ( i = 1 \vee P(B_{i-1,l}|X_l) < \beta_i)  
\end{alignat*}
\begin{equation}
\label{eq:MultiClassOneSkew}  
P(M|X,\Phi) = P(M_{i,l}|X_l) \sim Multinomial(\Phi_i) 
\end{equation}
\vspace{-0.5cm}
\begin{alignat*}{2}
    \text{where } & P(M_{i,l}|X_l) \ge \Phi_i \ \wedge  &\ (i = 1 \vee P(M_{i-1,l}|X_l) < \Phi_i)  
\end{alignat*}
\begin{equation}
\label{eq:mainOneSkew}
\begin{split}
Y_l &= \argmax{c}(P(Y\mid X_l,M_{i,l},\Phi_i,B_{i,l},\beta_i)) \\
    &= \argmax{c} \left(  
    \begin{dcases}
        P(B|X,\beta) &if \ P(B_{i,\zeta})>P(B_{i,\xi})\\
        P(M|X,\Phi) & \ otherwise
      \end{dcases} \right) 
\end{split}
\end{equation}

In this work, we use the variables, e.g.: $B$ and $M$, to represent both the probability distributions of the multistage classifiers and the classifiers. 
The dashed arrow in Figure \ref{fig:GM_OneSkewedClass} represents prevalence of the distribution of the parent variable over the child when a given a priori condition occurs $(P(B_{i,\zeta})>P(B_{i,\xi}))$. $\zeta$ represents the index of the majority class from the Bernoulli distribution generated by the binary multistage classifier B. Similarly, $\xi$ is the index of the minority classes cluster from the same Bernoulli distribution.  

\section{\textit{CMC-M}: \textit{C}o-Multistage of \textit{M}ultiple \textit{C}lassifiers for \textit{M}ultiple Skewed Classes}
\label{section:CMC-M}
Co-Multistage of Multiple Classifiers for Multiple skewed classes (\textit{CMC-M}) is a new stochastic model to support datasets with multiple majority-skewed classes. \textit{CMC-M} is an extended version of the \textit{CMC} model, introduced in Section \ref{section:CMC}.

\begin{figure} 
\centering
\includegraphics[width=0.5\textwidth]{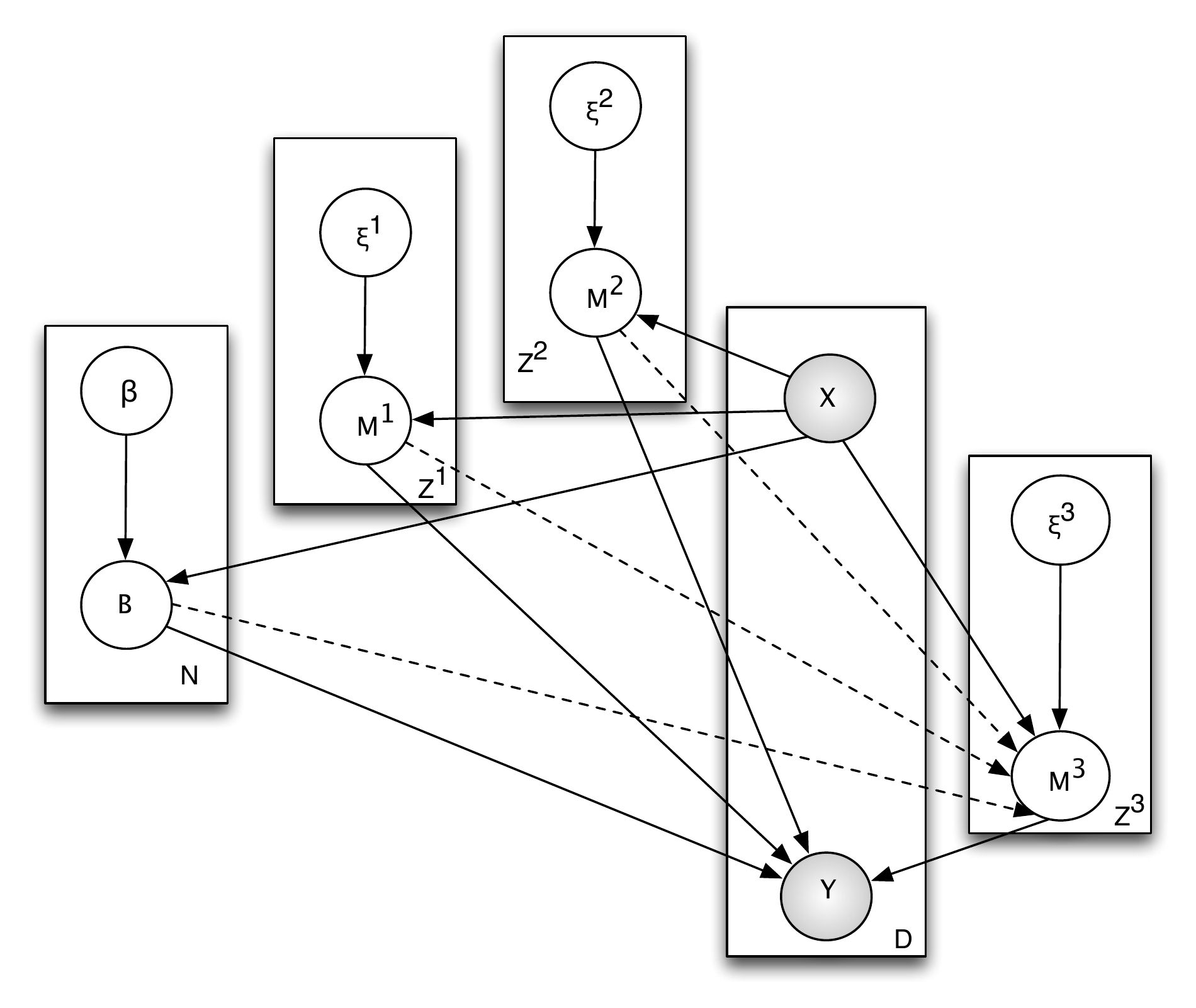}
\caption{Graphical model of the \textit{CMC-M}. The boxes represent replicates in the plates notation.
The dashed arrows are a extension to the plates notations.  It represents the prevalence of the parent variable distribution over the child under a priori condition.}
\label{fig:GM_MultipleSkewedClasses}
\end{figure}

As Figure \ref{fig:GM_MultipleSkewedClasses} shows, \textit{CMC-M} is made of two layers containing four multistage ensemble of multiple classifiers. Both layers have 
latent variables ($\beta$, $\xi^t$) over classifiers to determine their activation in the multistage. The main differences between \textit{CMC} and the \textit{CMC-M} frameworks are the inclusion of two extra multistage classifiers on the top layer and the activation of bottom layer multistage classifier. The activation of the bottom layer classifier ($M^3$) occurs when there is a disagreement in the quorum of the 3 top layer classifiers ($B$, $M^1$, $M^2$). More precisely when their probability distributions do not agree on the allocation of more probability mass to either a majority or minority class (Eq. \ref{eq:mainSeveralSkew}).

The top layer requires three transformations of the label space. The first label transformation is to adapt for the binary classifier, that classifies an instance into majority or minority class. We cluster the labels into majority class and minority class clusters. Because the majority classes are skewed, they can be trivially identified by comparing with the expected optimal balance distribution for each class ($\frac{|D|}{|\bar{Y}|}$ where D is the set of instances and Y the set of labels).

In addition, the two multistage multiclass classifiers with reduced label space help in the decision. While one classifier generates a probability distribution for the majority class cluster and each minority class, the other classifier generates a probability distribution for the minority class cluster and each majority class.
At the bottom layer is the default multistage multi-class classifier without changes in the label space (all labels). The bottom layer classifier distribution is used to recover from disagreements at the top layer.

Formally, the \textit{CMC-M} can be viewed as a generative process for each label Y in a set of instances D described by a set of features X:
 \begin{enumerate}
 \item Choose $\beta_i\ \in\ ]0,1]$, where $i\ \in\ \{1,...,N\}$ and N is the number of stages in the binary classification of majority cluster vs. minority cluster. The default value for $\beta_i$ is 1. 
 
 \item Choose $\xi_i^t \ \in\ ]0,1]$, where $i\ \in\ \{1,...,Z^j\}$,  $Z^j$ is the number of stages in the multiclass classification, and $j$ index the type multiclass classifier. Repeat the process for all $j \in {1,2,3}$,. When $t$ equals one, $\xi_i^1$ are variables over the multiclass classification of the majority classes cluster versus each minority class. $j$ equal to two defines the multiclass classification of the minority classes cluster vs each majority class. Finally, $j$ equals to three shows the multiclass classification with all classes. 
 
 \item For each label $Y_l$, where $l\ \in\ \{1,...,D\}$ and D is the number of instances.
 	 \begin{enumerate}
	 \item Choose $B_{i,l}$ based on Equation \ref{eq:bin}; (b) Choose $M_{i,l}^t$, for all $t \in \{1, 2, 3\}$ based on Equation \ref{eq:defaultMultiSkew}; (c) Choose label $Y_{l}$ according to Equation \ref{eq:mainSeveralSkew}.
	 \end{enumerate}
 \end{enumerate}
 \begin{equation}
 P(M^t|X,\phi)=P(M_{i,l}^t|X) \sim \ Multinomial(\xi_i^t) 
 \label{eq:defaultMultiSkew}
 \end{equation}
 \begin{alignat*}{2}
     \text{where } & P(M_{i,l}^t|X_l) \ge \xi_i^t \ \wedge &  & ( i = 1 \vee P(M_{i-1,l}|X_l) < \xi_i^t) 
 \end{alignat*}
 \begin{equation}
 \begin{split}
 Y_l &= \argmax{c} (P(Y|X,M^t,\xi^t,B,\beta)) 
     = \argmax{c} \left( 
  \begin{dcases}
     P(M^1|X,\xi^1) &if \ P(B_{i,\zeta})>P(B_{i,\phi}) \\ & \wedge  P(M^1_{i,\gamma})>P(M^2_{i,\Omega})\\
     P(M^2|X,\xi^2) &if \ P(B_{i,\zeta})<P(B_{i,\phi}) \\ & \wedge  P(M^1_{i,\gamma})<P(M^2_{i,\Omega})\\
     P(M^3|X,\xi^3) & \ otherwise
 \end{dcases}
 \right)
 \end{split}
 \label{eq:mainSeveralSkew}
 \end{equation}

Again, to simplify the notation, we used the variables $B$, $M^1$, $M^2$, $M^3$, to represent the probability distributions of both the multistage classifiers and the classifiers distributions. In addition, in Fig. \ref{fig:GM_MultipleSkewedClasses} the dashed arrows represent the prevalence of distribution of the parent variable over the child under the conditions described in Eq. \ref{eq:mainSeveralSkew}. $\zeta$ represents the index of the majority class from the Bernoulli distribution generated by the binary multistage classifier $B$. Similarly $\phi$ is the index of the minority classes cluster from the same Bernoulli distribution. While, $\gamma$ and $\Omega$ are the indexes of the majority and minority class cluster distribution generated by the multinomial multistage classifiers $M^1$ and $M^2$.

\section{Experimental Setup}
\label{sec:Experimental_Setup}
The classification results of the various algorithms are calculated using 
80\% of the dataset for training and 20\% for testing.
The initial evaluation included two datasets where the state-of-art cost sensitive method AdaC2 was performed.  
However the characteristics of the 2 UCI datasets did not allow to extensively stress our methods in the presence of a higher number of classes, namely skewed majority classes. For this purpose, we added three extra datasets from text classification tasks.

Both \textit{CMC} and \textit{CMC-M} have multistage ensemble of multiple individual classifiers. All classifier follow an architecture made of three classifiers with default parameters values, except the first. 
The sequence of classifiers in the multistage ensembles follows one heuristic. It starts with the simplest/weakest classifier and continues till the most complex/strongest. The current order was determined experiments by evaluating each heterogeneous classifier individually. The classifiers heterogeneity also improves imbalanced classification~\cite{Hsu:2009}.   

The first classifier is a Random Forest (RF) 
with 100 trees. The second corresponds to the SMO SVM 
with polynomial kernel and logistic probabilistic models. The last classifier is the ensemble of the previous two classifiers. It selects the classifier with higher probability or confidence value. 
There are several reasons that justify choosing both RF and SVM as base multistage classifiers. In the literature, it is not rare to find SVMs as the best performing supervised classifiers (specially in large feature spaces, e.g.: $>$ 1000 features per instances). Furthermore, they explore the whole feature space by introducing support vectors that describe the hyperplanes. Such hyperplanes divide the data by maximizing the distance between examples of different classes. Conversely, the RF selects random subsets of features to avoid the overfitting problem.  
Random feature selections is the simplest feature selection method~\cite{Motoda:2003,Tseng:2008,Hong:2009,Bailey:2013}. 
The runtime complexity of both architectures is in the worst case the complexity of the SMO classifier $O(L \cdot n)$, where is $n$ is the training sample size and L is the average number of (candidate) support vectors.

\subsection{Datasets}
\label{subsection:datasets}

We started our evaluation with two publicly available datasets at the UCI ML Repository
to compare our work with the state-of-art~\cite{sun2006boosting}.
The \textit{Car Evaluation} dataset is one of the most popular datasets in the UCI ML Repository. It contains 1728 instances described by 6 nominal ordered attributes. 
The \textit{New-Thyroid} dataset was built to predict patients' thyroid level class. This corresponds to three classes: euthyroidism (normal), hypothyroidism, and hyperthyroidism. This is a small dataset with 215 examples of patients, each described by five attributes. The normal class corresponds to the most frequent class, with 69.77\%. 
But these datasets hardly captures the several skewed majority classes problem.
As a result, we included 4 extra datasets from text classification task. Data sets \textit{fbis}, \textit{tr21}, and \textit{news3} are derived from the \textit{TREC-5/6/7}. The features included were stemmed word counts and stopword words were removed. 
Finally, we built our own large dataset using ACE 2005 dataset and 27 events classes\footnote{\url{http://to.disclose.after.review.process}} to show the weakness of G-mean and propose alternatives. This dataset has 1 very skew class, no event, and several classes with few instances per class ($<$ 50 instances).

\subsection{Results}
\label{subsection:results}
First, we evaluated \textit{CMC} on two UCI datasets to compare with state-of-art methods~\cite{sun2006boosting}: Adaboost.M1 over decision tree C4.5 
as weak classifier and AdaC2.M1 over C4.5 (cost-sensitive baseline). 
The cost setups of \textit{AdaC2} (short for AdaC2.M1) are predicted by a Genetic Algorithm. The results of C4.5 were also included to complement the baseline information. In addition to the cost-sensitive baseline, we also report the combination of our new \textit{CMC} and \textit{CMC-M} methods with preprocessing methods, as they benefit some classifier, e.g.: SVM~\cite{Aijun:2006,Batista:2004}. 
For this purpose, we selected all minority classes and oversampled them using \textit{SMOTE} with the default proposed configuration of 5 nearest neighbors and a new instance per each minority classes instance.
This means that we increased by a factor of two the number of minority classes instances. 
Then, we undersampled the dataset to 90\% of the original size with bias to a balanced distribution by randomly subsampling with replacement.   
In general the undersampling to 90\% of the original size yields better results than other resampling percentages 
 in our preliminary experiments with one skewed class. Even so, 90\% is suboptimal value when there is 2 or more skewed class, e.g.: undersampling to 79\% of the \textit{fbis} dataset yielded better results than the performance of \textit{CMC-M} with 90\% undersampling (
F1-Macro: 0.621, \textit{G-Mean}: 0.626). However, fine tuning the undersampling \textit{G-mean} metric~\cite{kubat1997addressing} yielded very small improvements. 
The first dataset used in our evaluation is the \textit{Car evaluation}. It has one majority class (unacceptable), which is the condition necessary to apply our \textit{CMC} model. 
Table \ref{tab:resultsOnCarsNewThyroid} reports the experiments on Car dataset.  The combination of undersampling with \textit{CMC}, (\textit{CMC (U.)}) improved \textit{G-Mean} by 10.3\% points, while the \textit{SMOTE} lowers the results by 2\%. 
The interpretation of why SMOTE lowers the results in some datasets is on the decision region for the minority class that can actually become smaller and more specific as the number of minority samples increases. But at the same time it is also prone to learn incorrect boundaries when the new synthetic samples are incorrectly labeled as minority instances. Furthermore, \textit{G-mean} of \textit{CMC (U.)} is 5.1\% points higher than the best baseline model \textit{AdaC2} with cost vector [0.541, 0.823, 0754, 1.000]. Also, the smallest minority class (very good) is fully identified without any error (F1 = 100\%).
\begin{table}[!htbp]
  \centering
  \caption{Results on Cars, News-Thyroid evaluation datasets}
  
    \renewcommand{\tabcolsep}{0.07cm}
    \begin{tabular}{r|r|rrr|rrrr}
    \toprule
    
            &       & \multicolumn{3}{|c|}{\textbf{Baselines}} &       &       &       &    \\
    Measure & Dataset & C4.5 & Ada.M1 	&	AdaC2		   & CMC (U.)  & CMC (O.) & CMC (O.U.) & CMC \\ \hline
    Macro-F1 &  Cars     & 0.829 & 0.877 & 0.911 &\textbf{0.934} & 0.869 & 0.911 & 0.880 \\
    G-Mean &       & 0.834 & 0.876 & 0.915 &\textbf{0.966} 	  & 0.843 & 0.959 & 0.863 \\ \hline
    Macro-F1 & New-   & 0.884 & 0.879 & 0.905 &\textbf{0.978} & 0.917 & 0.941 & 0.941 \\
    G-Mean &  Thyroid    & 0.854 & 0.866 & 0.901 &\textbf{0.969} & 0.923 & 0.956 & 0.956 \\
    \bottomrule
    \end{tabular}%
  \label{tab:resultsOnCarsNewThyroid}%
\end{table}%
The \textit{New-Thyroid} dataset has less one class than \textit{Cars} and one skewed majority class (normal). 
Table \ref{tab:resultsOnCarsNewThyroid} shows that undersampling improved \textit{CMC} \textit{G-Mean} values by + 1.3\% and 6.8\%. While oversampled \textit{CMC (O.)} reduces the G-mean results by 3.3\%. However, despite the reduction of performance cause by SMOTE oversampling, these results are still above the best baseline (+ 2.1\%) \textit{AdaC2} (cost vector: [0.421, 0.626, 1.000]). 
The rationale for the results of \textit{CMC (O.U.)} (CMC with SMOTE oversampling and undersampling) and \textit{CMC} (without sampling preprocessing) being about the same is justified by two reasons. First, it is the small size of the data that limits the application of the oversampling leading to the generation of bad data points and limiting the undersampling gains.
Consequently, we opted to analyze the performance of \textit{CMC-M} in two additional datasets from TREC evaluation. These datasets have two or more skewed majority classes, a larger number of classes and features. In addition, we included, a third dataset with one majority class, \textit{tr21}) to analyze the performance of \textit{CMC} with more classes and a larger feature space.

For the remaining datasets Adabost.M1 was selected as baseline method. Both \textit{AdaC2} and C4.5 were replaced with the results of the two classifiers used by the multistage classifier. This reduces redundancy of including C4.5 results and the unavailability of standard \textit{AdaC2} implementation.

After comparing the results on Tables \ref{tab:resultsOnTr21}, \ref{tab:resultsOnCarsNewThyroid}, we find that \textit{CMC (U.)} has again the best overall performance. The tr21 is another dataset with one imbalanced class (class label 251 corresponds to 68.8\% of the dataset). Both the Random Forest (RF) and the Adaboost.M1 baselines failed to identify one class. As a result, \textit{G-Mean} values were equal to 0.000. 
Another interesting results drawn from Table \ref{tab:resultsOnTr21} are the absolute improvements of F1 (+5.9\%) of \textit{CMC (O.U.)} over \textit{CMC}. These results are justified by improvements of precision of some of the minority classes. These improvements are obtained at the expense of misclassifying and reducing the recall of other less frequent minority classes as the lower \textit{G-Mean} (-4.6\%) indicates.   
\begin{table}[!htbp]
  \centering
  \caption{Results on tr21 dataset}
    \renewcommand{\tabcolsep}{0.07cm}
    \begin{tabular}{r|lrl|rrrr}
    \toprule
            &      \multicolumn{3}{|c|}{\textbf{Baselines}} &       &       &       &    \\
    Measure & RF & SMO & Ada.M1 						    & CMC (U.)  & CMC (O.) & CMC (O.U.) & CMC \\ \hline
    \midrule
    Macro-F1 & 0.173 & \textbf{0.736} & 0.282 & 0.663 & 0.655 & 0.690 & 0.631 \\
    G-Mean & 0.000 & 0.706 & 0.000 & \textbf{0.798} & 0.795 & 0.729 & 0.775 \\
    \bottomrule
    \end{tabular}%
  \label{tab:resultsOnTr21}%
\end{table}%
The following tables \ref{tab:resultsonFbis} and \ref{tab:resultsNew3Dataset} were generated using 2 datasets with multiple skewed classes to enabled a better understanding of both \textit{CMC-M} and \textit{CMC} methods. 

With the 4 majority classes and 13 minority classes, it was expectable that \textit{CMC-M} outperformed \textit{CMC} on \textit{fbis} as observed. To clarify this assumption, we shall note that only three majority classes are skewed, with a representation of 20.5\%, 15.7\%, and 14.5\% of the \textit{fbis} dataset. The fourth majority class is slightly skewed with representation of 7.7\% of the \textit{fbis} dataset and the average class representation is 5.9\%. The slightly skewness of the fourth majority class is not taken into account by our methods. 

The data sparsity is frequently found in supervised learning tasks with large number of features, such as text classification. The \textit{fbis} dataset with a feature set made of 2000 is a example of a text classification sparse dataset. Naturally, oversampling methods are more useful in the presence of data sparsity because they will fill empty spaces. Otherwise, those classification spaces would be empty and lead to misclassification. 
At the same time, undersampling of the majority classes in large feature spaces can reinforces the data sparsity because undersampling of the majority classes explores the removal of redundancy which might not not exist in small/medium datasets (e.g.: $<$ 1,000,000 instances). The consequence is the reduction of the classifier performance, unless the classes are very easily separable and the dataset distribution becomes balanced. Based on these assumptions, it is not surprising to observe that \textit{CMC-M (O.)} outperforms \textit{CMC-M} with regard to \textit{G-Mean} improving it 0.6\%. 
The results of \textit{CMC-M (U.)} and \textit{CMC-M (O.U.)} confirmed again the previous assumptions about the prejudicial effect of undersampling to datasets with large number of classes and features. The poor performance of\textit{ CMC-M (O.U.)} might be justified by the combination of two side effects of both oversampling and undersampling. First is the inclusion of some new incorrectly classified minority classes instances, and removal of known and non-redundant majority class instances, mislead the classifiers into identifying less precise decision boundaries. 
\vspace{-0.5cm}
\begin{table} 
  \centering
  \caption{Results on fbis dataset}
    \renewcommand{\tabcolsep}{0.06cm}
    \begin{tabular}{r|rrr|r|llll}
    \toprule
            & \multicolumn{3}{|c|}{\textbf{Baselines}} &       &       &       &  &  \\
    Measure & RF & SMO & Ada. & CMC & CMC-M& CMC-M & CMC-M & CMC-M\\
    	    &    &     & M1   &(O.) & (U.)& (O.) & (O.U.)  &\\
    \midrule
    Macro-F1& 0.586 & 0.648 & 0.535 & \textbf{0.738} & 0.617 & 0.703 & 0.590 & 0.697\\
    G-Mean & 0.000 & 0.574 & 0.000 & 0.652 & 0.624 & \textbf{0.678} &  0.599 & 0.674\\
    \bottomrule
    \end{tabular}%
  \label{tab:resultsonFbis}%
\end{table}%
\vspace{-0.5cm}
The last dataset used in our experiments was \textit{news3}. With nearly 10,000 instances and 27,000 features has five times more instances and about fourteen times more features than the \textit{fbis} dataset. At the same time, the number of classes is 44, that makes it the dataset with largest number of classes used in our evaluation. 
From the 44 classes, we identified 15 majority classes(2.3\% of the dataset). 
But, only 3 classes of the 15 majority classes are above for example twice the average number of instances per class with 5.9\%, 7.3\%, and 5.1\% of the dataset. Thus, we believe that it would be useful to have a more fine-grained distinction of classes distribution than just two types: majority and minority class found in the literature allow. For example, a ternary categorization made of majority ($|y_i|> 2\times \bar{Y}$), minority ($|y_i|< \frac{1}{4}\bar{Y}$), intermediate ($\frac{1}{4}\bar{Y} <|Y_i|< 2\times \bar{Y}$) classes. 
Still, we assumed the 15 most frequent classes as majority classes and the remaining as minority classes in our experiments using the \textit{news3}. Table \ref{tab:resultsNew3Dataset} shows the results of those experiments.  Those results confirmed that undersampling reduced \textit{CMC-M} overall performance by 4.4\%. At the same time, we also observed another expectable results about oversampling. As the number of features, instances, and classes grows, it is hard to generate new points because they need to be at the same time relevant, correctly labeled, and should not collide with other classes data points. However, these properties are not guaranteed by oversampling techniques, such as \textit{SMOTE}. Therefore, we did not observe improvements of applying oversampling before \textit{CMC-M} in this dataset. Indeed, the \textit{G-mean} differences between \textit{CMC-M (O.)} and \textit{CMC-M} were found not to be statistically significant. 
 However, both versions of \textit{CMC-M} are statistically significative better than the baselines.
\begin{table} 
  \centering
  \caption{Results on the new3 dataset}
    \renewcommand{\tabcolsep}{0.07cm}
    \begin{tabular}{r|lrl|rrrr}
    \toprule
            &      \multicolumn{3}{|c|}{\textbf{Baselines}}      &       &       &  &  \\
    Measure & RF & SMO & Ada.M1& CMC-M (U.)& CMC-M (O.)& CMC-M (O.U.)& CMC-M\\
    \midrule
    Macro-F1& 0.689 & 0.700 & 0.594 & 0.656 & 0.712 & 0.648 & \textbf{0.713} \\
    G-Mean & 0.590 & 0.649 & 0.498 & 0.657 & 0.698 & 0.640 & \textbf{0.701} \\
    \bottomrule
    \end{tabular}%
  \label{tab:resultsNew3Dataset}%
\end{table}%
\vspace{-1.2cm}
\begin{table} 
  \centering
  \caption{ACE 2005 dataset with 27 events}
    \begin{tabular}{r|rr|rrrrr}
    \toprule
		    & \multicolumn{2}{c|}{\textbf{Baselines}}    &  &    &   &   &    \\
    Measure & RF & SMO & CMC(U.)& CMC(O.)& CMC(O.U.)& CMC  & CMC-M \\
    \midrule
    Macro-F1& 0.055 & 0.300   & \textbf{0.346} & 0.300   & 0.344 & 0.308 & 0.297 \\
    G-Mean & 0     & 0     & 0     & 0     & 0     & 0     & 0     \\
    SG-Mean & 0.002 & 0.060 & 0.305 & 0.164 & \textbf{0.366} & 0.202 & 0.146  \\
    \# $R_i$=0 & 25    & 8     & 2     & 3     & \textbf{1} & 2     & 3   \\
    \bottomrule
    \end{tabular}%
  \label{tab:ACE-2005-27-events-20K-features}%
\end{table}%
\vspace{-0.3cm}
The \textit{CMC} improves the recall on the ACE dataset at the expense of losing some precision at the detection of the majority classe(s) - in this case the no-event. But \textit{CMC (O.U.)} enabled us to detect 26 event classes, where the baseline methods only detect 2 and 19 . 
Translating to $SG\text{-}Mean = \left(\prod_{i=1}^{n} R_i + \delta \right)^{\frac{1}{n}} , \ \delta > 0$  - smooth G-Mean with $\delta = 0.001$) improvement of about 183 and 6 times more.   
%
\section{Conclusions}
\label{section:conclusions}
In this paper, we presented 2 new stochastic models to increase the robustness of imbalanced multiclass learning. These 2 models explore different classes topologies.
The first model, \textit{CMC}, improves imbalanced multiclass learning with one skew majority class state-of-art.
The literature does not provide a clear distinction between majority and minority classes for imbalanced multiclass classification. One of the possible reasons for this gap is in focus on the binary case as 3 recent surveys about imbalanced learning~\cite{galar2012review,he:2009learning,sun2009classification} claim. Another justification is that most research works  
focused on datasets with few classes, where a precise distinction between majority classes and minority classes is not necessary. Thus, in this work, we formally defined the distinction between majority class ($|y_i|> \bar{Y}$) and minority class ($|y_i| \le \bar{Y}$) 
as a class whose number of instances is higher or lower than the average
Another contribution of this work is the creation of imbalanced multiclass learning models for datasets having small and large number of both classes and features. 
For example, the  
\textit{CMC-M} 
 was designed for imbalanced multiclass learning of large number of classes.

The overall absolute improvements (\textit{G-Mean}) of the \textit{CMC} combined with undersampling over the best baseline (\textit{AdaC2}) ranged between +5\% and +7\%. 
A small percentage of undersampling 
 based on random majority class selection bias towards the balanced distribution with replacement, clearly helps the \textit{CMC} model. Caused by 
the reduction of the imbalanced distribution towards the unique majority class. However, the undersampling effect is not always beneficial. For instance, it reduces the \textit{CMC-M} performance in the presence of several majority classes.
This reduction occurs because undersampling of the majority classes in large feature spaces can reinforces the data sparsity because undersampling of the majority classes explores the removal of redundancy which might not not exist in small/medium datasets. 
The consequence is the reduction of the classifier performance, unless the classes are very easily separable and the dataset distribution becomes balanced. The reduction of performance of the classifier is also observable internally in the reduction of agreement between binary classifier and two top layer multi-class classifiers. 

Finally, \textit{SMOTE} improves \textit{CMC-M} results for small datasets, such as \textit{fbis}, but it should not be used for large datasets, e.g.: \textit{new3}, unless there are classes with very few instances e.g.: ACE 2005. Namely for datasets with classes with few instances, it is possible to observe a recall values equal to zero. 
Therefore, we have also proposed a new evaluation metric \textit{SG-Mean}. 
 
In future work, we will investigate the expansion of the \textit{CMC-M} architecture to include a ternary representation of classes distribution.
%
%
%
%
%
\bibliographystyle{IEEEtran}
%
\bibliography{bare_conf}

\end{document}